\documentclass[pdflatex,sn-mathphys-num]{sn-jnl}% Math and Physical Sciences Numbered Reference Style
\usepackage{graphicx}%
\usepackage{multirow}%
\usepackage{amsmath,amssymb,amsfonts}%
\usepackage{amsthm}%
\usepackage{mathrsfs}%
\usepackage[title]{appendix}%
\usepackage{xcolor}%
\usepackage{textcomp}%
\usepackage{manyfoot}%
\usepackage{booktabs}%
\usepackage{algorithm}%
\usepackage{algorithmicx}%
\usepackage{algpseudocode}%
\usepackage{listings}%
\theoremstyle{thmstyleone}%
\theoremstyle{thmstyletwo}%

\theoremstyle{thmstylethree}%

\usepackage{array}
\begin{document}

\title[Article Title]{Modular Multi-Task Learning for Chemical Reaction Prediction}

%%=============================================================%%
%% GivenName	-> \fnm{Joergen W.}
%% Particle	-> \spfx{van der} -> surname prefix
%% FamilyName	-> \sur{Ploeg}
%% Suffix	-> \sfx{IV}
%% \author*[1,2]{\fnm{Joergen W.} \spfx{van der} \sur{Ploeg} 
%%  \sfx{IV}}\email{iauthor@gmail.com}
%%=============================================================%%

\author*[1]{\fnm{Jiayun}\sur{Pang}\email{j.pang@gre.ac.uk}}

\author[1]{\fnm{Ahmed M.}\sur{Zaitoun}}

\author[1]{\fnm{Xacobe Couso}\sur{Cambeiro}}

\author*[2]{\fnm{Ivan}\sur{Vulić}\email{iv250@cam.ac.uk}}

\affil*[1]{\orgdiv{School of Science, Faculty of Engineering and Science}, \orgname{University of Greenwich}, \orgaddress{\street{Medway Campus, Central Avenue}, \city{Chatham Maritime}, \postcode{ME4 4TB}, \country{United Kingdom}}}

\affil*[2]{\orgdiv{Language Technology Lab}, \orgname{University of Cambridge}, \orgaddress{\street{9 West Road}, \city{Cambridge}, \postcode{CB3 9DA}, \country{United Kingdom}}}

%%==================================%%
%% Sample for unstructured abstract %%
%%==================================%%

\abstract{Adapting large language models (LLMs) trained on broad organic chemistry to smaller, domain-specific reaction datasets is a key challenge in chemical and pharmaceutical R\&D. Effective specialisation requires learning new reaction knowledge while preserving general chemical understanding across related tasks. Here, we evaluate Low-Rank Adaptation (LoRA) as a parameter-efficient alternative to full fine-tuning for organic reaction prediction on limited, complex datasets. Using USPTO reaction classes and challenging C–H functionalisation reactions, we benchmark forward reaction prediction, retrosynthesis and reagent prediction. LoRA achieves accuracy comparable to full fine-tuning while effectively mitigating catastrophic forgetting and better preserving multi-task performance. Both fine-tuning approaches generalise beyond training distributions, producing plausible alternative solvent predictions. Notably, C–H functionalisation fine-tuning reveals that LoRA and full fine-tuning encode subtly different reactivity patterns, suggesting more effective reaction-specific adaptation with LoRA. As LLMs continue to scale, our results highlight the practicality of modular, parameter-efficient fine-tuning strategies for their flexible deployment for chemistry applications.}


%%\pacs[JEL Classification]{D8, H51}

%%\pacs[MSC Classification]{35A01, 65L10, 65L12, 65L20, 65L70}

\maketitle

\section{Introduction}\label{sec1}

Large language models (LLMs) represent a groundbreaking development in AI, with far-reaching impact across multiple fields. In chemistry and related scientific disciplines, LLMs are increasingly applied over challenges such as molecular property prediction, novel molecular structure generation, reaction outcome prediction and retrosynthesis planning\cite{White_LLMs_review}. These applications generally follow two complementary directions. One direction involves augmenting LLM with chemistry-specific toolkit packages, such as ChemCrow\cite{ChemCrow,10.1093/bib/bbaf601}, which integrates expert-designed chemistry tools into an LLM agent for synthesis planning and materials discovery. The second direction uses LLMs for downstream tasks via transfer learning. In this scenario, pre-trained LLMs (often trained on massive amounts of raw text and/or general chemistry knowledge in terms of molecules and reactions)\cite{Schwaller:2019molecular,nach0,Chemformer,Jablonka2024,Zhao2025,Choi2025} are fine-tuned for new, domain specific tasks using comparatively small, labeled datasets. For example, several recent studies, including our own,\cite{D4FD00104D} demonstrate that finetuned language models, such as FlanT5\cite{flant5-2} and ByT5\cite{byt5}, deliver reasonable performance on chemical reaction prediction and optimisation tasks.\cite{qiao2022transformer,EmmaKingTransferLearn, Kotlyarov2024, Pesciullesi2020,sagawa2023reactiont5, Wieczorek2025}

Fine-tuning is a powerful technique for adapting LLMs to domain-specific tasks. However, this transfer learning paradigm poses considerable challenges\cite{10.1145/3735633}. First, conventional fine-tuning, often referred to as full fine-tuning, requires updating all model parameters to fit the new task, which can be computationally and memory intensive. Second, adapting to new tasks may lead to catastrophic forgetting of the knowledge acquired during pre-training - when the fine-tuned model overfits the typically limited new-task data. This can lead to poor performance on the original tasks, making it difficult to achieve good performance across multiple tasks.\cite{MCCLOSKEY1989109,goodfellow2015,li2024revisiting_cf_llm} To address these issues, parameter efficient fine-tuning (PEFT) techniques have emerged in recent years\cite{Ding2023}. These approaches enable tuning large models without updating all the weights and parameters. Prompt- and prefix-based PEFT methods\cite{lester-etal-2021-power,li2021prefixtuning} emphasise light-weight and task-specific adaptation, while modular methods such as Low-Rank Adaptation (LoRA) offer reusability and have become widely adopted\cite{houlsby2019parameter, pfeiffer2020adapterhub, hu2021lora, wang2024peft_survey,xin-etal-2024-beyond}. A LLM contains many weight matrices ($W$) that map an input vector of dimension \textit{d} to an output vector of dimension \textit{k}, and $W = d \times k$. In full fine-tuning, all entries in these large matrices are updated, i.e. $d \times k = 4096 \times 4096 =$ 16.8 million parameters per matrix in a GPT-3 style model, which is computationally expensive and requires substantial data to avoid overfitting. With LoRA, instead of retraining the full weight matrix $W$, it keeps $W$ frozen and introduces two trainable low-rank matrices $A$ and $B$ with a small rank $r$ (e.g. 4, 8 or 16), where $A = r \times d$ and $B = k \times r$. This means that $\Delta W = BA = 4 \times (4096 + 4096) = 32768$ parameters to learn, only 0.2\% of the original 16.8 millions parameters. Now the LoRA-updated weight matrix $W’$ is given as 

$W' = W + \Delta W = W + BA$

Where $W$ is frozen and the lightweight learned adjustment $\Delta W = BA$ captures task-specific information. Because only the small matrices A and B are trained, LoRA dramatically reduces the number of trainable parameters, often to under 1\% of the model, thus significantly lowering computational costs while achieving performances comparable to full fine-tuning. More importantly, all task-specific new learning is contained entirely within the small matrices A and B (referred to as a LoRA module), which act as a self-contained adaptation module that captures knowledge specific to a particular task or dataset. These modules are independent of the base model and can be added or removed without retraining the base model. In addition, one base model can support multiple LoRA modules - switching tasks/datasets then means switching LoRA modules, not retraining the entire model, thus avoids interference between tasks. By preserving most of the initial parameters, LoRA helps prevent catastrophic forgetting by maintaining general knowledge embedded in the base model. 

In organic reaction prediction, forward reaction, retrosynthesis and reagent predictions are three fundamental yet inter-related tasks. They can therefore be leveraged to build more robust multi-task models, which benefit from shared representations and cross-task learning. Recent studies, such as the T5-based models\cite{Lu:2022unified, D4FD00104D} have demonstrated that multi-task learning enhances reaction prediction by improving data efficiency and enabling generalisation without the need to focus on a single, pre-defined task. Typically, the multi-task reaction prediction models are trained on broad organic chemistry datasets, such as USPTO\cite{Lowe2017}, allowing them to capture extensive general knowledge of reaction pattens. In practice, researchers must often further fine-tune such general models on smaller, domain-specific datasets containing more complex or novel reaction types, which are the typical datasets in chemical and pharmaceutical R\&D. In this context, it is essential that the fine-tuned model not only learns new, specialised knowledge but also retains its broader chemical understanding to remain effective across related tasks. Conventional full fine-tuning often fails to achieve this balance.\cite{Wieczorek2025,Sivaraman} For example, it has been observed that fine-tuning a multi-task model for a single task produces only marginal improvement in reagent prediction accuracy (Acc@1 from 24.15 to 24.38), but resulted in catastrophic forgetting of the forward reaction task (Acc@1 drops from 95.82 to 35.26) due to excessive task-specific adaptation (i.e. overfitting).\cite{D4FD00104D} The findings highlight the limitations of full fine-tuning for multi-task chemical models. 

\begin{figure}[tbp]
\centering
\includegraphics[width=\textwidth]{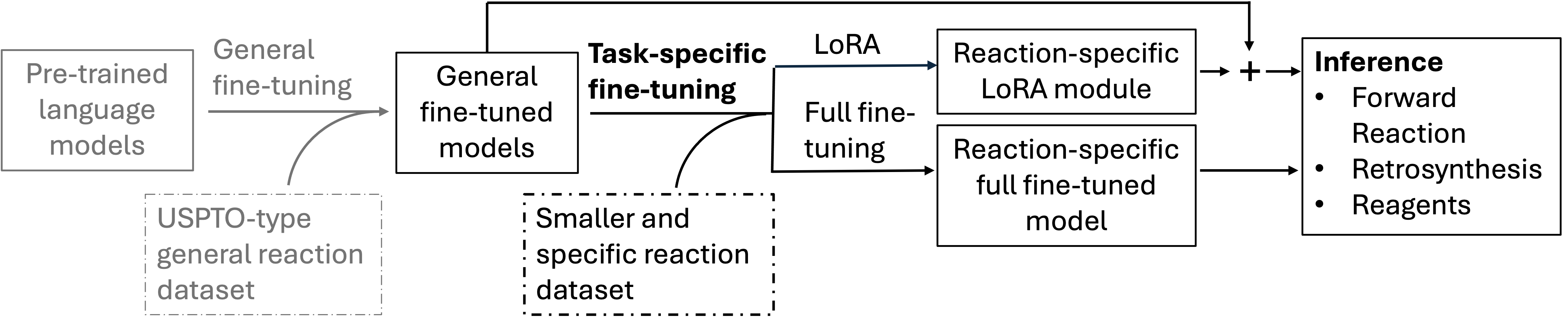}
\caption{The framework of our approach: general full fine-tuning from a pre-trained language model using the USPTO\_1K\_TPL dataset, followed by task-specific adaptation via either full fine-tuning and LoRA using a smaller and specific reaction dataset (e.g. C-H functionalisation). The general fine-tuning step has been reported in our previous work (in grey) while the current study focuses on the task-specific finetuning step. The model at each stage is multi-task in nature and can be fine-tuned either in a single task or multi-task fashion.}\label{fig1}
\end{figure}

In the present work, we investigate LoRA’s capability to mitigate overfitting and catastrophic forgetting when specialising general trained multi-task chemical reaction prediction models on smaller and specific reaction datasets (Figure~\ref{fig1}). Our approach leverages LoRA’s modular architecture in which weight updates for each reaction type are encapsulated within a lightweight LoRA module, while the parameters of the large base model remain frozen during fine-tuning. The reaction type-specific modules can then be dynamically combined with the general-purpose knowledge embedded in the base model, enabling flexible and efficient adaptation to new reaction types. We benchmarked this modular strategy across a range of general organic chemistry reaction classes from the USPTO\_1K\_TPL dataset\cite{Schwaller2021} and the more challenging metal-catalysed C-H functionalisation reaction, evaluating performance on forward reaction prediction, retrosynthesis and reagent prediction tasks. The results demonstrate that the modular LoRA approach delivers accuracy comparable to full fine-tuning for individual tasks, while offering a substantial advantage in maintaining accuracy across multiple tasks – highlighting its robustness in multi-task settings. Moreover, both LoRA and full fine-tuned models exhibit generative capabilities and can predict chemically plausible alternative solvents that are not present in the training data. Notably, fine-tuning on C-H functionalisation reactions shows that the models capture slightly different chemical reactivities in their representation space, indicating effective adaptation to reaction-specific features in LoRA. Although we have used the older generation of T5-styled models as the base models for this proof-of-principle study, the modular solutions are equally applicable to the latest generation of LLMs. Given the rapidly increasing scale and complexity of state-of-the-art LLMs, our work highlights the importance of adopting modular and parameter-efficient fine-tuning approaches to enable flexible and knowledge-preserving adaptation for diverse applications in chemistry.

\section{Results}\label{sec2}
\subsection{Accuracy of full fine-tuning vs LoRA benchmarked using ten random classes from USPTO\_1K\_TPL.}\label{subsec2.1}

To validate the LoRA approach, we first benchmark it against full fine-tuning using ten randomly selected reaction classes from the USPTO\_1K\_TPL dataset (Further information of the reaction classes can be found \url{https://rxn4chemistry.github.io/rxnfp/label_template_visualisation/}). We compare the accuracy of prediction in three models: direct evaluation using the general fine-tuned model (without task-specific adaptation), full fine-tuning and LoRA. Each model is evaluated using the three standard reaction prediction tasks - reagent prediction, forward reaction prediction and retrosynthesis - using base models ByT5 small, ByT5 base\cite{byt5} and nach0 base.\cite{nach0} nach0 is a T5-style language model pretrained jointly on language and chemical corpora to form a shared representation space. It is included in the initial evaluation stage to assess whether such hybrid pretraining provides a more advantage initialisation points for LoRA finetuning.

\begin{figure}[ht]
\centering
\includegraphics[width=\textwidth]{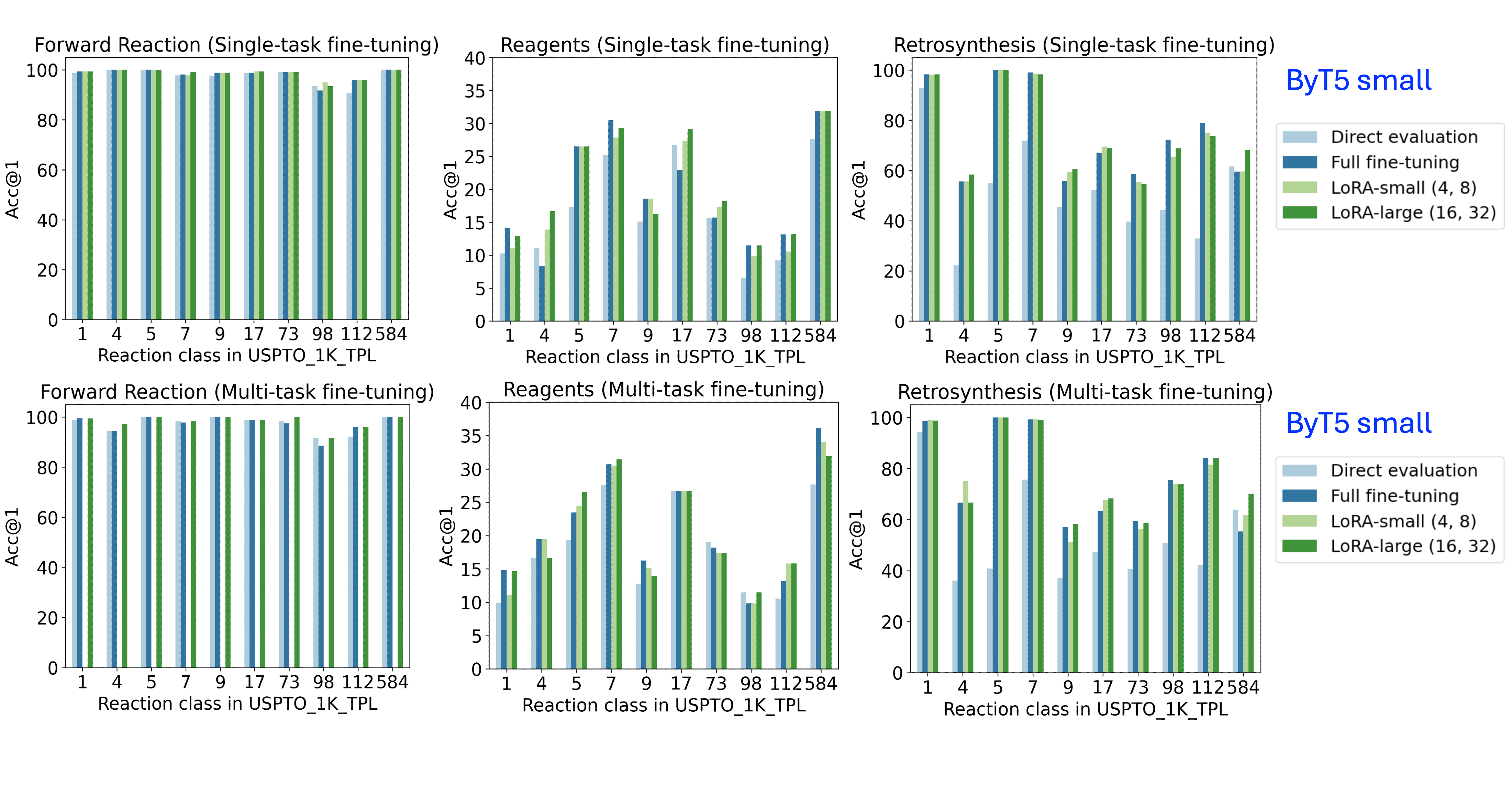}
\caption{Results in the three evaluation tasks with three different models based on ByT5 small: the general fine-tuned model (without task-specific adaptation), full fine-tuning and LoRA with two sets of parameters. LoRA-small uses r=4 and $\alpha$=8 while LoRA-large uses r=16 and $\alpha$=32.}\label{fig2}
\end{figure}

\begin{figure}[htbp]
\centering
\includegraphics[width=\textwidth]{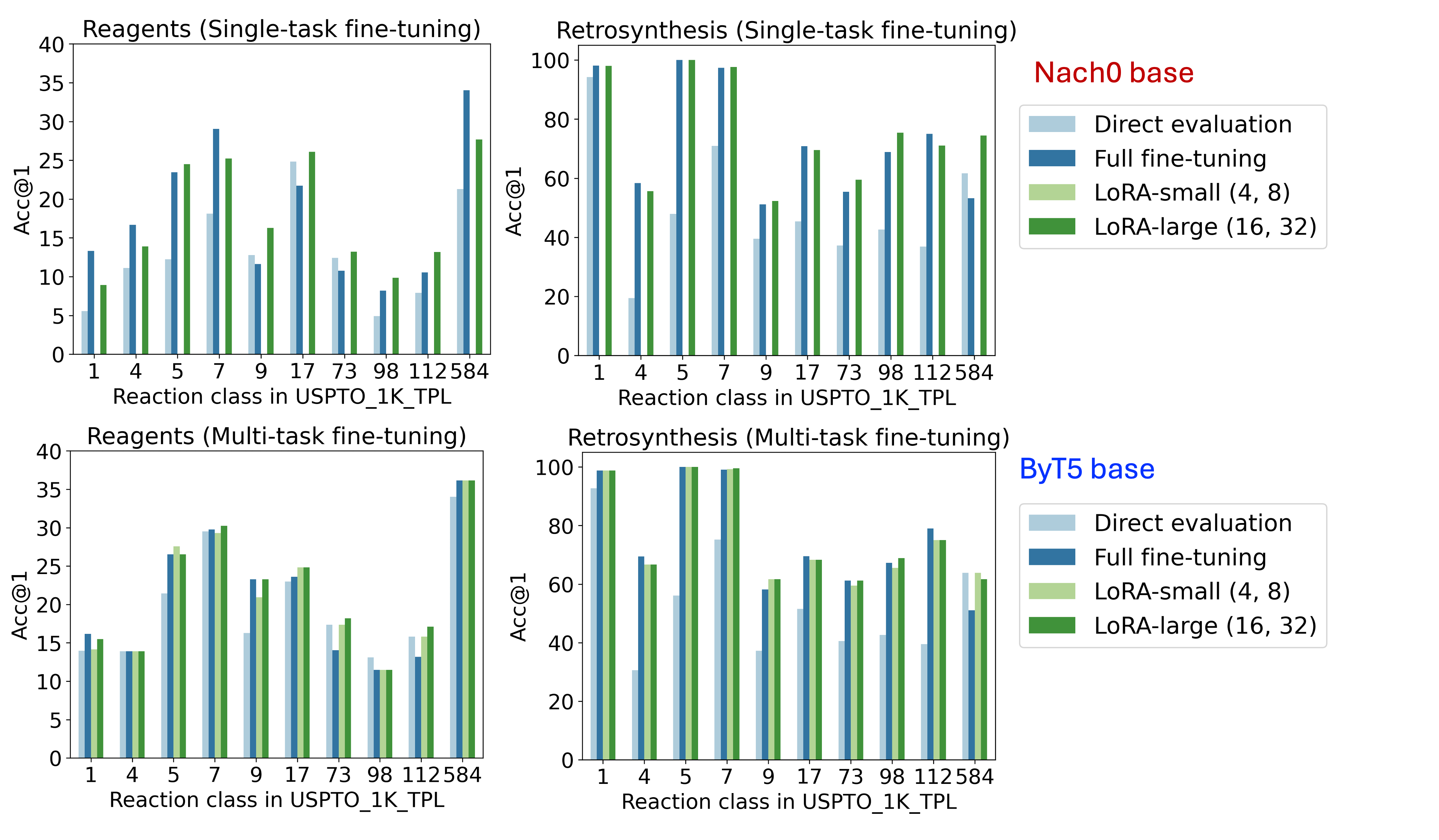}
\caption{Results in tasks of reagents prediction and retrosynthesis with three different models based on nach0 base (top) and ByT5 base (bottom): the general fine-tuned model (without task-specific adaptation), full fine-tuning and LoRA with one or two sets of parameters. LoRA-small uses r=4 and $\alpha$=8 while LoRA-large uses r=16 and $\alpha$=32.}\label{fig3}
\end{figure}

Substantial variation is observed across individual reaction classes but as expected, both full fine-tuning and LoRA improve prediction accuracy across tasks relative to direct evaluation (Figure~\ref{fig2}). Forward reactions are predicted with extremely high accuracy ($>$90\%) by all models, likely due to the strong one-to-one correspondence between reactant–product pairs (Note: these reactions were classified using a template extract method\cite{ThakkarTemplate} which only considers the immediate neighbourhood of the reaction, therefore reactions in each class tend to share a highly similar reaction centre and are extensively represented in the training data). For reagents prediction and retrosynthesis, both full fine-tuning and LoRA give modest yet consistent improvements over direct evaluation, representing a medium effect according to Cliff’s delta. In addition, the Wilcoxon test indicates a statistically significant difference (p $<$ 0.05) across the ten reaction classes. It suggests that full fine-tuning and LoRA generally or always outperforms direct evaluation for retrosynthesis and reagent prediction for general organic chemistry reactions. 

Overall, LoRA on average achieves marginally higher average accuracy than full fine-tuning (by 0.3–2\%), with larger LoRA ranks (r) generally yielding a further improvement by a few percent. In some cases (e.g. class 584, in both single-task and multi-task fine-tuning with ByT5 small and Nach0 base, Figure~\ref{fig2} and Figure~\ref{fig3} respectively), a more substantial increase in accuracy is observed using LoRA. Nevertheless, the improvement introduced by LoRA over full fine-tuning is not statistically significant across the ten classes of reactions with both single-task and multi-task fine-tuning. This is consistent with other work indicating that LoRA performs close to full fine-tuning but whether LoRA can give better accuracy depends on multiple factors, such as the dataset size and complexity and the base model architecture.\cite{biderman2024lora,gravereaux2025}

\subsection{Comparison of chemical space coverage between full fine-tuning and LoRA}\label{subsec2.2}

\begin{figure}[htbp]
\centering
\includegraphics[width=\textwidth]{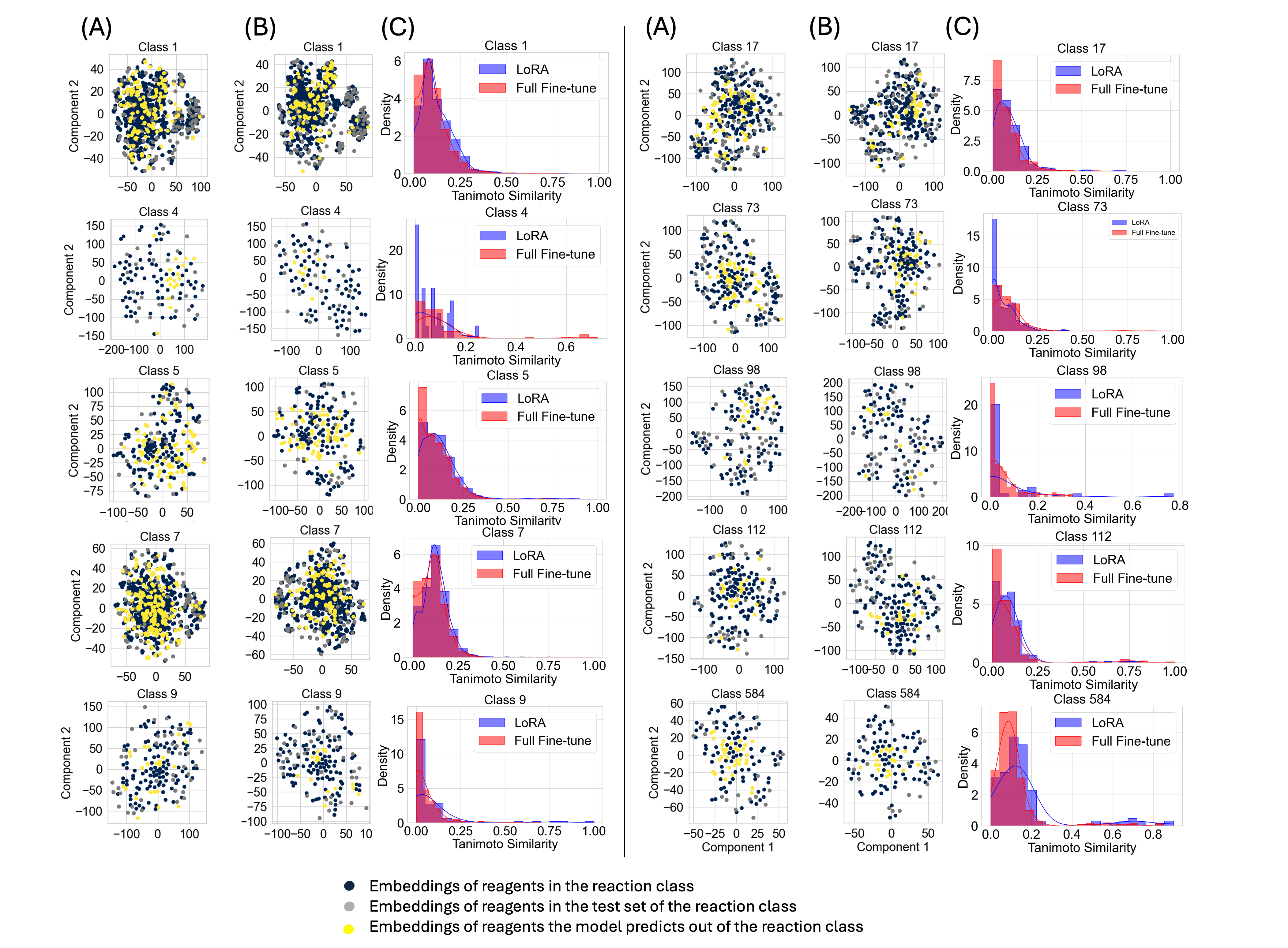}
\caption{t-SNE visualisation of the embeddings of out-of-distribution reagents (yellow dots) from (A) the full finetuned model and (B) the LoRA model. Blue dots represent the embeddings of all reagents in the corresponding reaction class, while grey dots represent reagents only in the test split for that class. (C) Tanimoto similarity distribution for the out-of-distribution reagents proposed by the LoRA model (blue column) and the fully fine-tuned model (red column).}\label{fig4}
\end{figure}

When training machine learning models for reaction prediction, a key challenge is not merely reproducing known chemistry but generalising to new reagents, conditions, and reaction contexts that the model has not explicitly seen during training.\cite{D2SC06798F,Toniato2023} Traditional feature-based models, which depend on predefined molecular descriptors, often perform well at pattern recognition within the training distribution but typically struggle to extrapolate beyond it. In contrast, by learning contextual and relational patterns at multiple levels, LLMs have been shown to be able to propose chemically plausible predictions not explicitly observed during training.\cite{Jablonka2024,cha-lee-2024-evaluating} We use the reagent prediction task to examine the models’ ability to generalise because reagents, such as solvents have complex roles in chemical reactions and the relationship between reagents and reaction outcomes is context dependent. Across the ten USPTO\_1K\_TPL reaction classes, both fully fine-tuned and LoRA models can propose reagent molecules (within the top 5 predictions) that fall outside the reaction-class-specific fine-tuning set and the USPTO\_1K\_TPL set. These out-of-distribution reagents appear scattered in embedding space: some are loosely clustered (e.g. class 4 and 584), while most show no clear pattern. Overall, full fine-tuning generates a larger number and greater diversity of out-of-distribution reagents than LoRA, as reflected in the broader Tanimoto similarity distribution (where 0 denotes no similarity and 1 denotes identical molecules, Figure~\ref{fig4}). 

\subsection{A multi-task fine-tuning approach for C-H borylation reactions}\label{subsec2.3}

\subsubsection{Product prediction of the C-H dataset}\label{subsubsec2.3.1}

To more rigorously evaluate the performance of LoRA fine-tuning, it is important to test it on reactions that are more challenging than the general organic reactions included in the USPTO dataset. For this purpose, we select C-H functionalisation reactions, specifically the iridium-catalysed C–H borylation as the benchmark. C–H functionalisation reactions are difficult to predict because they often involve multiple, nearly equivalent reactive sites and their site- and regio-selectivity are governed by a delicate interplay of factors, such as catalyst, reagents and substrate structure \cite{D5SC00541H,Nippa2024,Lapkin}. This makes them an ideal benchmark set for assessing the model’s ability to distinguish between similar reactive sites in forward reaction prediction and to predict subtle reactivity variations in reagent prediction, which provides a strong indication of the model’s underlying chemical understanding and its transferability to novel reaction types. Our results show that direct evaluation from general finetuned model does not work at all (0.7 Acc@1), thus task-specific fine-tuning is essential to adapt to the reaction set (Table 1). Full fine-tuning with ByT5-small offers decent accuracy on the forward reaction prediction (in a single task fashion) (69.6 Acc@1). Task-specific LoRA fine-tuning can further improve the accuracy to 78.3 Acc@1. The larger ByT5-base model with the same set of LoRA parameters offers a marginal improvement to 79.7 Acc@1. The accuracies of our models are in the same range as previous studies.\cite{Nippa2024,Kotlyarov2024} 

\begin{table}[ht]
\centering
\textbf{Table 1:} Combinations of hyperparameters for the top 1, top 2, top 3 and top 5 accuracy of C-H borylation forward reaction prediction. The numbers in the brackets are LoRA r, a and dropout; \textit{e} stands for the number of epochs and \textit{lr} is the learning rate. 
\begin{tabular}{ l  l  l  l  l }
\toprule
\textbf{ByT5-small} & Acc@1 & Acc@2 & Acc@3 & Acc@5 \\
\midrule
-- No fine-tuning & 0.7 & 1.3 & 2.0 & 2.0 \\

Full fine-tuning  & 69.6 & 79.7 & 84.0 & 91.3 \\

LoRA (16, 32, 0.0), lr=0.001 & 72.5 & 89.9 & 91.3 & 94.2 \\

LoRA (16, 32, 0.0), lr=0.003 & 78.3 & 85.5 & 87.0 & 91.3 \\

LoRA (16, 32, 0.05), lr=0.003 & 75.4 & 87.0 & 91.3 & 95.6 \\

\textbf{ByT5-base} &  Acc@1 &  Acc@2 &  Acc@3 &  Acc@5 \\

LoRA (16, 32, 0.0), lr=0.003 & 79.7 & 88.4 & 92.8 & 94.2 \\

LoRA (16, 32, 0.05), lr=0.003 & 79.7 & 89.9 & 92.8 & 94.2 \\
\bottomrule

\end{tabular}
\label{table1}

\end{table}

\begin{figure}[htbp]
    \centering
    \includegraphics[width=\textwidth]{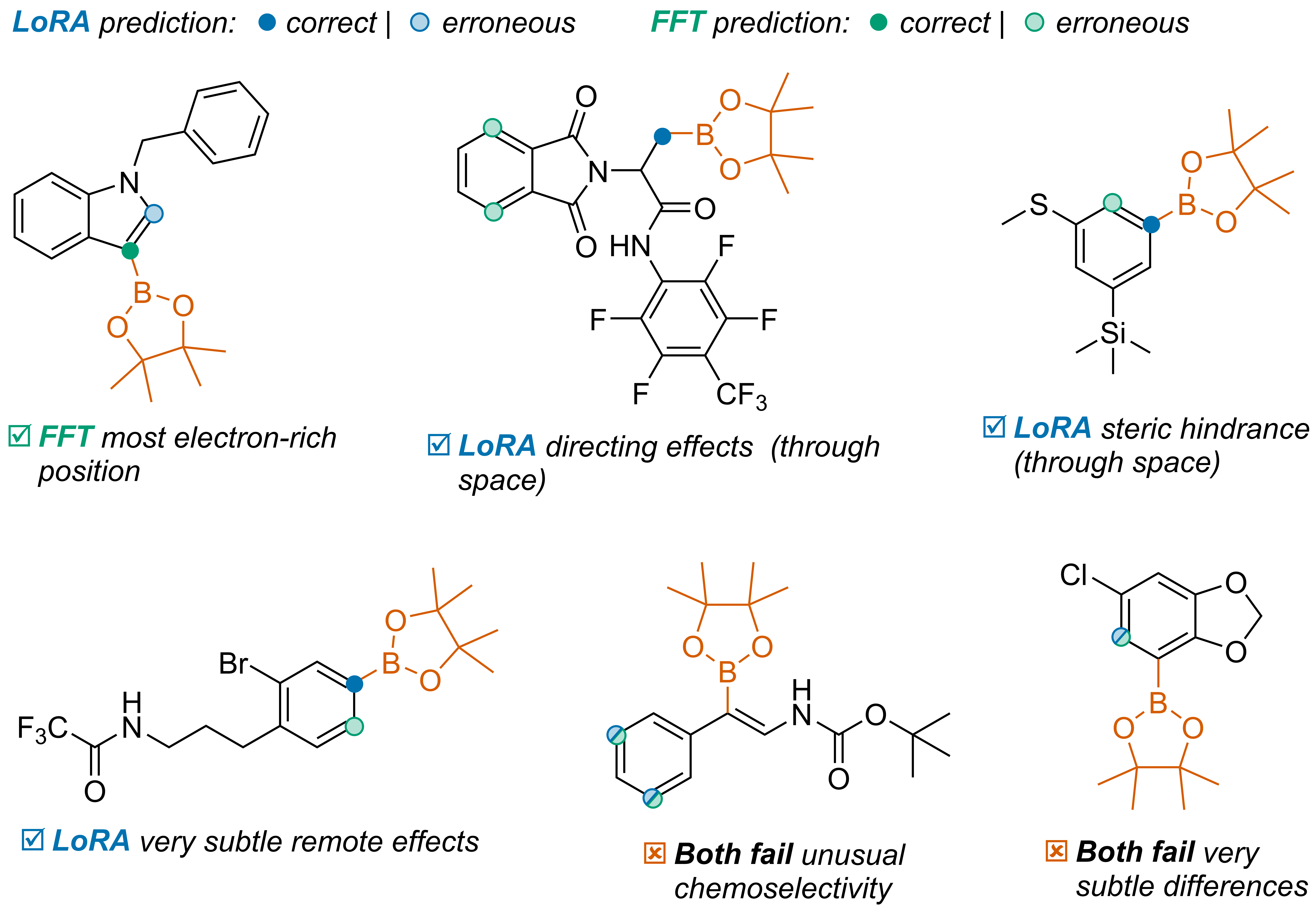}
    \caption{Comparison between the C-H borylation product prediction (Acc@1) using LoRA and full fine-tuning (ByT5 small).  Incorrectly predicted activation positions are highlighted in lighter blue circle (LoRA) and green circle (full fine-tuning/FFT), while correctly predicted position are shown in solid circles. The nature of selectivity for each product is also given.}
    \label{fig5}
\end{figure}

The accuracy difference between full fine-tune and LoRA corresponds to a few reactions that LoRA correctly predicts while the full fine-tuning fails (Figure~\ref{fig5}). We observe that the full fine-tuned model tends to favour electron-rich activation position, reflecting general chemical intuition whereas the LoRA model tends to predict activation position that may be more specific to the training dataset. This suggests that while both approaches capture subtle reactivity patterns, LoRA might be more effective at adapting to reaction-specific features, such as directing effect and steric hindrance that are important for this class of C-H activation reactions, as illustrated Figure~~\ref{fig5}. Both models fail at some reactions with unusual selectivity or with very subtle differences. These cases represent intrinsically challenging scenarios for data-driven models, as regioselectivity is influenced by subtle electronic and steric effects that are not always captured by molecular representations alone. Notably, such ambiguities are not limited to computational approaches; even experienced experimental chemists often require DFT calculations or experimental validation to identify the most reactive or electron-rich position. 

\subsubsection{Mitigation of catastrophic forgetting by LoRA}\label{subsubsec2.3.2}

\begin{figure}[htbp]
\centering
\includegraphics[width=0.5\textwidth]{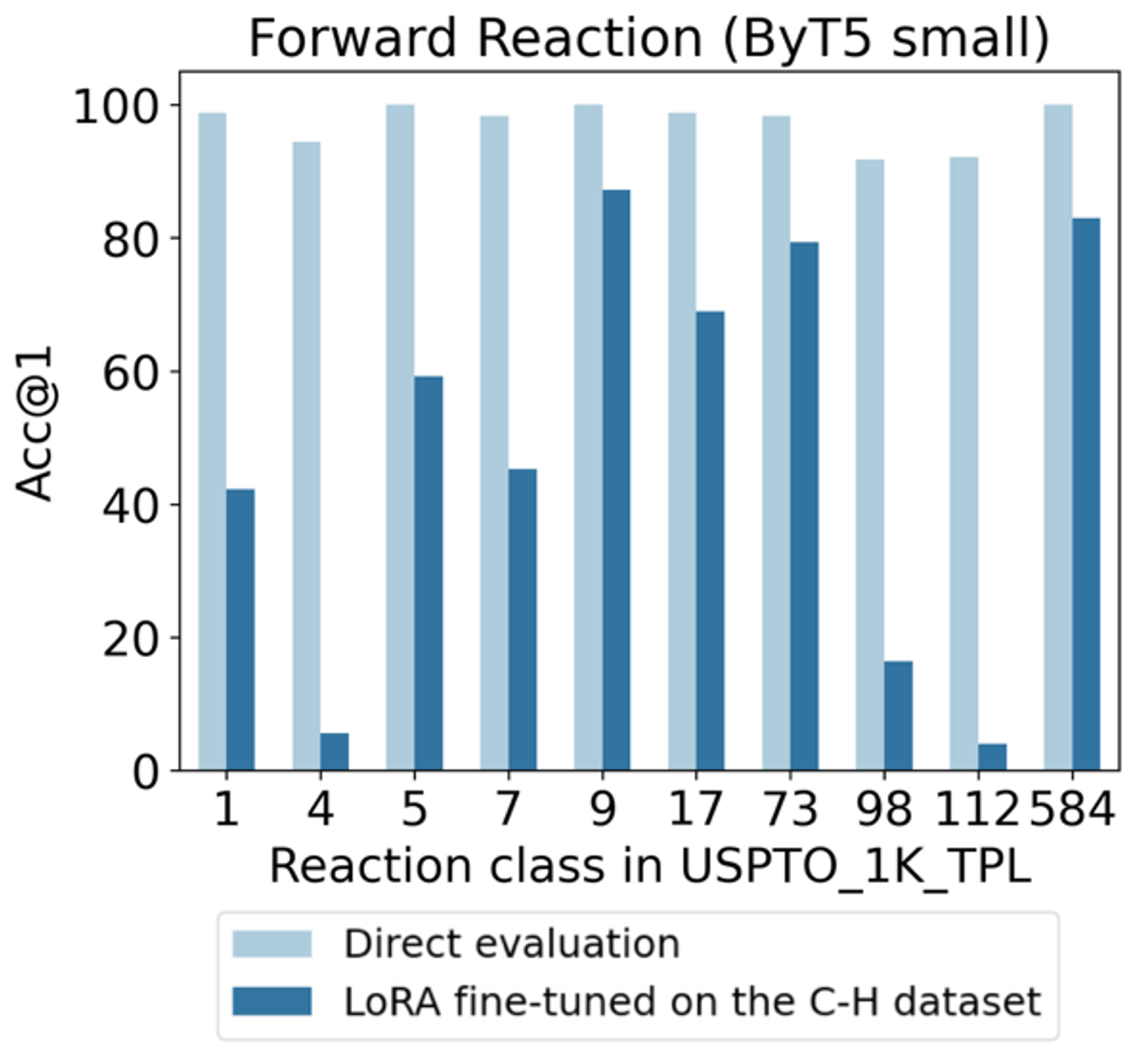}
\caption{Results of the forward reaction prediction on the ten classes of general organic reactions from USPTO\_1K\_TPL with two different models based on ByT5 small: the general fine-tuned model (direct evaluation, i.e. without task-specific adaptation) and the task-specific LoRA on the C-H borylation dataset finetuned in a multi-task fashion using forward reaction and reagent prediction. }\label{fig6}
\end{figure}

To evaluate LoRA’s ability to mitigate catastrophic forgetting, we tested the task-specific LoRA fine-tuned model (ByT5 small, multi-task finetuned on the C-H borylation dataset) on the ten classes from USPTO\_1K\_TPL (Figure~\ref{fig6}). As noted in Section 1, the general fine-tuned model achieves $>$90\% accuracy for forward reaction on these classes of general organic reactions. However, after further full fine-tuning using the C-H dataset, the model’s performance collapses, with  Acc@1 falling below 1\% across all ten classes, indicating severe overfitting to the task-specific chemistry. In contrast, the LoRA fine-tuned model retains a substantial amount of the original general reaction knowledge. Although performance decreases to varying degrees – most notably in classes 4, 98 and 112 – accuracy remains above 80\% for classes 9, 73 and 584 and between 40\%-65\% for the remaining classes. These results demonstrate that LoRA significantly reduces catastrophic forgetting by preserving the model’s broader chemical knowledge.

\subsubsection{Reagent prediction for the C-H dataset}\label{subsubsec2.3.3}

Both full fine-tuning and LoRA give excellent accuracy for reagent prediction of the C-H dataset. The 75-78 Acc@1 for reagent prediction is significantly higher than the accuracy ($<$40\%) over the USPTO\_1K\_TPL dataset (Figure~\ref{fig2} and Figure~\ref{fig3}), likely due to the smaller variation of reagents used in the C-H dataset. Furthermore, when task-specific fine-tuned in the multi-task fashion with both forward reaction and reagent prediction, LoRA was able to achieve excellent performance for both tasks at 75 Acc@1, while the full fine-tuning results in poorer performance of both tasks at 56 Acc@1 (Table 2). This is consistent with the general observation that learning signals from different tasks may negatively interfere with each other during finetuning and highlights the advantage LoRA may have in reducing interference when specialising general trained multi-task models for new type of reactions\cite{xin-etal-2024-beyond,zhang2025lori}.

\begin{table}[htbp]
\centering
\textbf{Table 2:} Task-specific fine-tuning for reagent prediction for the C-H dataset using both single-task and multi-task approach (LoRA parameters r = 16, $/alpha$ = 32 and dropout = 0.0; epoch=100 and lr = 0.01).
\begin{tabular}{ l  l  l  l  l  l  l }
\toprule
  &   &   & Acc@1 & Acc@2 & Acc@3 & Acc@5 \\
\midrule
\multirow{2}{*}{Single-task } & Full fine-tune & \multirow{2}{*}{REAG} & 78.3 & 85.5 & 87.0 & 87.0 \\

 & LoRA &  & 75.4 & 82.6 & 87.0 & 88.4 \\
\multirow{4}{*}{Multi-task} & \multirow{2}{*}{Full fine-tune} & FWD& 56.5 & 72.5 & 75.4 & 81.2 \\

 &  & REAG & 55.0 & 63.8 & 75.4 & 81.2 \\
 & \multirow{2}{*}{LoRA} & FWD & 75.4 & 88.4 & 88.4 & 91.3 \\
 &  & REAG & 74.0 & 81.2 & 87.0 & 88.4 \\ \bottomrule

\end{tabular}
\label{table2}

\end{table}

\begin{figure}[htbp]
\centering
\includegraphics[width=0.9\textwidth]{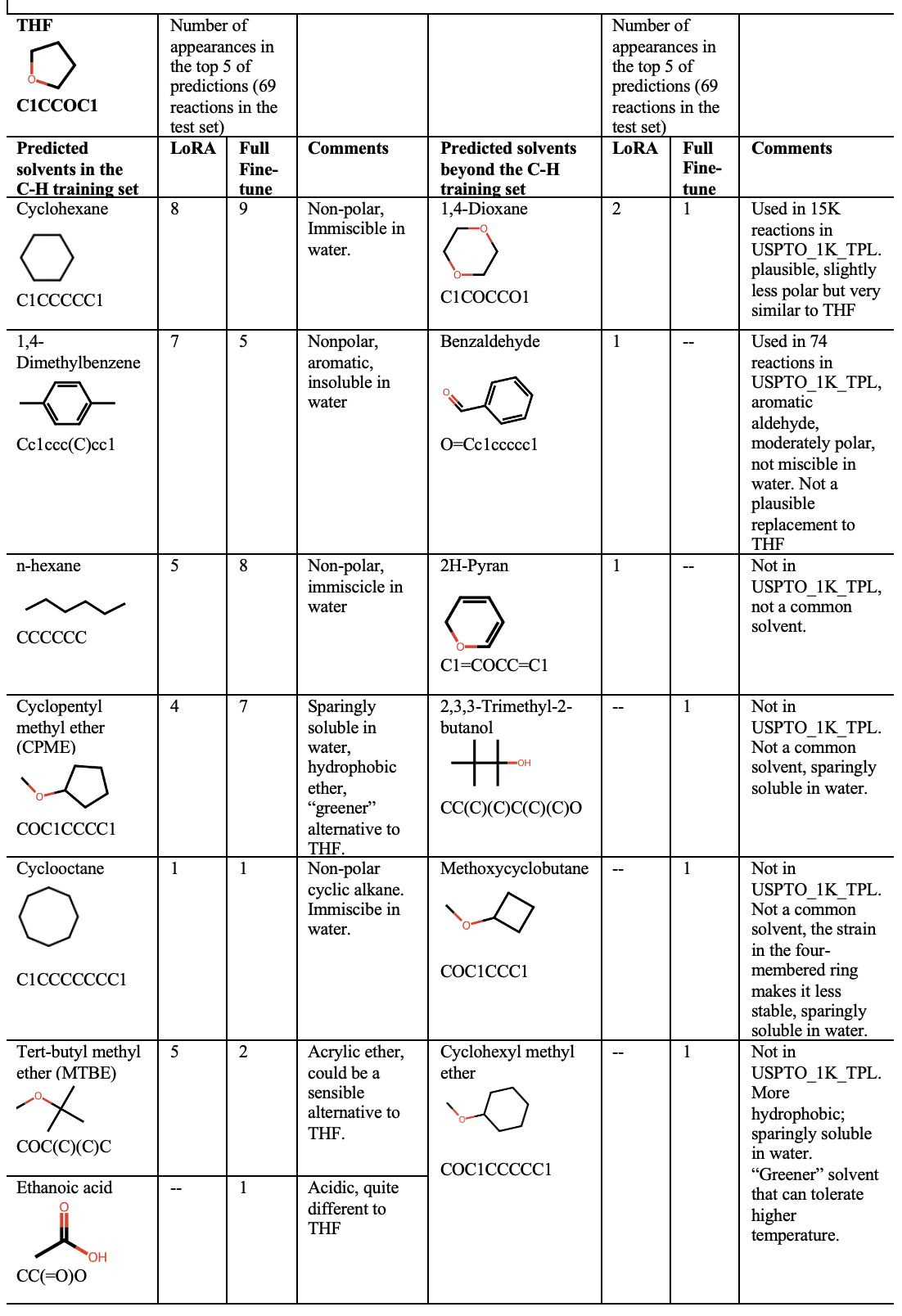}
\caption{Thirteen alternative solvents to tetrahydrofuran (THF) were predicted within the top-5 predictions of both the fully fine-tuned and LoRA models. The solvents in the left column are present in the C–H borylation dataset used for task-specific fine-tuning. The right column shows solvents drawn from broader sources: the top two appears in the USPTO\_1K\_TPL dataset used for general fine-tuning, whereas the bottom four do not appear in either dataset, indicating genuine extrapolative generalisation beyond the training distribution.}\label{fig7}
\end{figure}

We assess the C-H task-specific model’s ability to generalise beyond its training distribution, both beyond the C-H dataset and beyond the general fine-tuning dataset. Tetrahydrofuran (THF), one of the most common solvents in the C-H borylation reactions is used as an illustrative example (Figure~\ref{fig7}). THF is a polar, water-miscible ether, capable of dissolving a wide range of polar and non-polar substrates, making it a versatile general solvent. Both full fine-tuned and LoRA models propose solvents (within the top 5 predictions) that appear in the C-H borylation training dataset (left column, Figure~\ref{fig7}). Several of these, such as CPME and MTBE, are chemically reasonable substitutes based on polarity and structural similarity, while other less polar solvents could be plausible depending on reactant solubility.

Importantly, the models also generate six solvents that do not occur in the C-H borylation dataset (right column, Figure~\ref{fig7}). Of these, 1,4-dioxane and Benzaldehyde are in the USPTO\_1K\_TPL dataset used for general fine-tuning. 1,4-Dioxane, structurally similar to THF, represents a sensible alternative, whereas benzaldehyde is chemically distinct and unlikely to be a practical replacement. The remaining four predictions 2H-pyran, 2,3,3-trimethyl-2-butanol, methoxycyclobutane, and cyclohexyl methyl ether - appear in neither dataset, indicating genuine extrapolative generalisation. Several of these molecules reflect logical structural transformations of solvents in the training data (e.g., ring-size variations of CPME or ring expansion of THF). Overall, the results suggest that the task-specific models, whether trained via full fine-tuning or LoRA, can exhibit extrapolative generalisation and capture underlying chemical relationships in their representation space. Moreover, the fact that the two fine-tuning strategies sometimes propose different yet chemically plausible alternatives indicates that they may cover distinct regions of parameter space\cite{shuttleworth}, leading to complementary predictive behaviour.

\section{Methods}\label{sec3}

The general fine-tuning step has been reported in our previous work\cite{D4FD00104D} while the current study focuses on the task-specific finetuning (Figure~\ref{fig1}). We aim to evaluate the effectiveness of LoRA for further specialising general fine-tuned chemical reaction models on smaller, domain specific reaction datasets. We tested model architectures, datasets and LoRA hyperparameters across three typical organic reaction prediction tasks – forward reaction prediction, reagents prediction and retrosynthesis.

\subsection{Model architectures}\label{subsec3.1}
Our base model is the ByT5 pre-trained language models.\cite{byt5} Based on our previous work, performance differences between the small and base variants of ByT5 were minimal, therefore we primarily focus on ByT5 small (300M parameters) in this study. Following our established preprocessing pipeline, we trim the natural language sub-word vocabulary and retain only tokens that occur in SMILES sequences and their corresponding embeddings that occurred in SMILES sequences. For comparison, we also evaluated the nach0 base variant (220M parameters),\cite{nach0} a T5-style language model pretrained jointly on language and chemical corpora to form a shared representation space. We included nach0 in the initial evaluation stage to assess whether such hybrid pretraining provides a more advantage initialisation points for LoRA finetuning. All models follow the standard encoder–decoder structure of T5 where the input sequence (reaction SMILES) is fed to the model.

\subsection{Datasets}\label{subsec3.2}
We train and evaluate the models using two datasets. USPTO\_1K\_TPL\cite{Schwaller2021} contains 445k reactions divided into 1000 reaction templates serving as reaction class labels. It represents a diverse collection of general organic reactions and provides a strong basis for general fine-tuning. The C-H Borylation dataset contains borylation regioselectivity reactions as curated by Nippa et al from the literature\cite{Nippa2024,Kotlyarov2024}. The original dataset contains 1300 reactions. For this study, we filter entries according to the reaction yield ($>$0.3) and drop duplicate products, resulting in 685 reactions. The data were split into 8:1:1 for training (557 reactions), validation (69 reactions) and test (69 reactions), respectively.

\subsection{LoRA hyperparameters and training setup}\label{subsec3.3}
The ByT5 models are multi-task in nature. When fine-tuning, we define three tasks: forward reaction prediction, retrosynthesis, and reagent prediction. Each task is marked by a specific prefix (Product:, Reactants:, Reagents:), followed by SMILES of the different components of reaction arranged in the following format: 
Product: reactants.reagents>product
Reagents: reactants.product>reagents
Reactants: product>reactants
The general fine-tuned multi-task models can be further fine-tuned in a single task or multi-task fashion. When only the input sequence is provided without a specific prefix, then the model is task-specific fine-tuned for a single task. If two or three prefix are provided, the model is task-specific fine-tuned in a multi-task fashion. 

Due to the large number of experiments, we run a very constrained hyperparameter search. There may be slight changes in the results with a finer-grained hyperparameter optimisation but it is computational expensive and is beyond the time frame of the project. A few sets of LoRA parameters are explored, namely r, alpha and dropout along with the learning rate. The rank r controls the capacity of the LoRA adapter (typical values 4,8, and 16) – larger r indicates more parameters to update and smaller r indicates a lighter adaptation but may underfit a complex task. The scaling factor alpha control the magnitude of the LoRA update. Larger alpha means that scaling up the LoRA update to contribute more strongly to the model. Typical values of alpha are 16, 32 and 64. After some preliminary test, we used the r=16 and $\alpha$ =32 combination, which means the LoRA update is scaled by factor of 32/16=2. Dropout (i.e. a dropout layer on the input to the LoRA adaptor) is an optional parameter with typical values 0.05 or 0.1 (5\% or 10\% inputs are dropped at random) and is set to 0 for stable fine-tuning. Learning rates of 0.01 and 0.03 and 50-100 epochs were used. 

\subsection{Evaluation metrics}\label{subsec3.4}
The standard accuracy was reported at rank K (Acc@K) scores, measuring if the gold sequence can be found in the top K output sequences generated by the model. We select K-s following prior work (i.e., it is {1, 2, 3, 5}). To assess statistical significance of accuracy between the finetuning methods, statistical methods Cliff’s delta and the Wilcoxon signed-rank test were applied.

\section{Discussion and Conclusion}\label{sec4}

In summary, this study demonstrates that LoRA effectively enhance the performance of generally trained multitask chemical language models when specialised for new reaction types. Compared to full fine-tuning, LoRA's modular approach provides comparable or slightly improved accuracy while better preserving performance across multiple tasks within a model, such as the model’s ability for forward reaction and reagent prediction concurrently. More importantly, the modular nature of LoRA effectively mitigates catastrophic forgetting after task-specific finetuning, as indicated by maintaining approximately 50\% of accuracy averaged across ten reaction types compared to almost total loss of accuracy after full fine-tuning. Using the challenging C–H borylation as a benchmark, we show that LoRA and full fine-tuning models capture subtle reactivity patterns in the reactants and exhibit generative capabilities for reagent prediction indicative of underlying chemical understanding. The difference in predictions implies divergence in the learned parameter spaces of LoRA and full finetuning. 

This work builds on encoder–decoder T5-style models, which were among the most effective architectures at the time when we started the research. While newer generations of LLMs now offer more broader context, their ability to predict specific chemical reactions remain limited. For example, GPT-5 from OpenAI performed poorly on Suzuki coupling and Buchwald-Hartwig reactions under zero-shot inference, achieving accuracy lower than all fine-tuned models reported by Han et al.\cite{han2025} Task-specific fine-tuning is essential, as Han et al further demonstrate that LLaMa-3.1-8B, when finetuning on a small amount of specific reactions, achieves a slightly higher prediction accuracy than the chemistry-specific model Mol-T5, which follows a pre-training and fine-tuning strategy similar to that used in this work.\cite{han2025}  

Future work will focus on the latest open-source LLMs and expanding fine-tuning to a broader range of research-relevant reaction types to enable more systematic interpretation of how LoRA and full fine-tuned models may learn chemical reactivity differently. Ultimately, LoRA’s modular setup enables a highly flexible reaction prediction framework, where the base model provides general chemical understanding and individual LoRA module supply specialised knowledge for different reaction types or subtasks (e.g., forward reaction and reagent prediction are typical connected prediction need). By dynamically loading relevant LoRA modules, the framework introduces efficiency, generalisation, and chemical diversity and can be adapted to multi-reaction scenarios for data-driven discovery in organic synthesis.

\backmatter

%\bmhead{Supplementary information}

%Please refer to Journal-level guidance for any specific requirements.

\bmhead{Acknowledgments}

The work was supported by the UK Engineering and Physical Sciences Research Council (EPSRC) (grant number EP/Y004167/1) and a University of Greenwich Vice Chancellor PhD studentship to Ahmed M. Zaitoun. Ivan Vulić acknowledges the support of a Royal Society University Research Fellowship (no. 221137).

\bibliography{sn-bibliography}% common bib file
%% if required, the content of .bbl file can be included here once bbl is generated
%%\input sn-article.bbl

\end{document}